\documentclass[letterpaper, 10 pt, conference]{ieeeconf}  

\IEEEoverridecommandlockouts                              

\usepackage{bm}
\usepackage{amsmath} 

\usepackage{graphicx}

\usepackage{pgfplots}

\usepgfplotslibrary{fillbetween,groupplots,dateplot}
\usetikzlibrary{patterns,shapes.arrows}
\pgfplotsset{
	max space between ticks=30pt,
	try min ticks=3,
	every axis/.style={
		axis y line=left,
		axis x line=bottom,
		axis line style={thick,->,>=latex, shorten >=-.4cm}
	},
	every axis plot/.append style={thick},
	tick style={black, thick},
	compat=newest
}
\tikzset{
	semithick/.style={line width=1.1pt},
}
\title{\LARGE \bf
Can Transformer Attention Spread Give Insights Into Uncertainty of Detected and Tracked Objects?
}

\author{Felicia Ruppel$^{1,2}$, Florian Faion$^1$, Claudius Gl\"{a}ser$^1$ and Klaus Dietmayer$^2$
	\thanks{$^{1}$Robert Bosch GmbH, Corporate Research, 71272 Renningen, Germany, 
		{\tt\small \{firstname.lastname\}@de.bosch.com}}%
	\thanks{$^{2}$Institute of Measurement, Control and Microtechnology, Ulm University, Germany,
		{\tt\small \{firstname.lastname\}@uni-ulm.de}}%
}

\makeatletter
\let\NAT@parse\undefined
\makeatother
\usepackage{hyperref}

\begin{document}

\maketitle
\thispagestyle{empty}
\pagestyle{empty}

\begin{abstract}
Transformers have recently been utilized to perform object detection and tracking in the context of autonomous driving. One unique characteristic of these models is that attention weights are computed in each forward pass, giving insights into the model's interior, in particular, which part of the input data it deemed interesting for the given task. Such an attention matrix with the input grid is available for each detected (or tracked) object in every transformer decoder layer. In this work, we investigate the distribution of these attention weights: How do they change through the decoder layers and through the lifetime of a track? Can they be used to infer additional information about an object, such as a detection uncertainty?  Especially in unstructured environments, or environments that were not common during training, a reliable measure of detection uncertainty is crucial to decide whether the system can still be trusted or not. 

\end{abstract}

\section{INTRODUCTION}
Object detection and tracking are essential tasks in a perception pipeline for autonomous and automated driving. Only with knowledge about surrounding objects, downstream tasks, such as prediction and planning, are possible. In such a system, where the cascading effects of perception errors can be detrimental, it is very important to be able to quantify the reliability of the detection and tracking output. In object detection, uncertainty can stem from two sources \cite{feng2018uncertainty}: \textit{Epistemic} uncertainty is caused by uncertainty of the model, e.g. when an observation is made that was not present in the training dataset. Unstructured and dynamic environments can also cause such an uncertainty, as their versatility can hardly be captured in a training dataset. Second, \textit{aleatoric} uncertainty stems from sensor noise, and also encompasses uncertainty caused by low visibility and increased distance from the sensor \cite{feng2018uncertainty}.

While state-of-the-art object detection methods have been based on deep learning for many years, both with image input \cite{Girshick_2015_ICCV} as well as on point clouds \cite{lang_pointpillars:_2019, qi_pointnet++_2017}, it is a recent phenomenon that deep learning based models are also used for joint tracking and detection  \cite{luo_fast_2018,meinhardt_trackformer:_2021, ruppel2022trans}. Such trackers aim to utilize the detector's latent space to infer additional information about a tracked object, rather than relying on low-dimensional bounding boxes as input. However, they have the drawback that they are unable to output an uncertainty, as a conventional method would, e.g. tracking based on a Kalman filter \cite{chiu_probabilistic_2020}. While deep learning based detectors and trackers usually output a confidence score or class probability score per estimated object, these generally can not be used as a reliable uncertainty measure, but additional measures are necessary to capture uncertainty \cite{loquercio2020uncertainty}.
\begin{figure}[t]
	\includegraphics[width=\columnwidth]{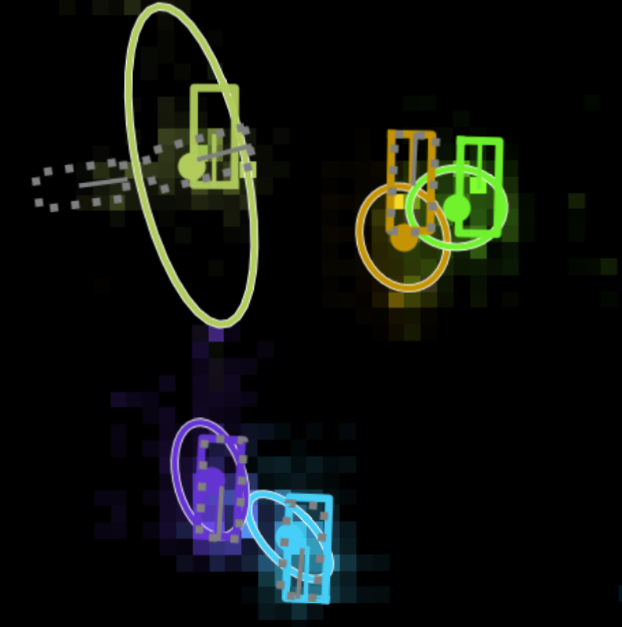}
	\caption{Example of estimated bounding boxes with their respective attention covariance matrices, pictured as ellipses. Ground truth boxes are denoted by dotted grey lines, while estimated boxes, attention weights, attention mean and ellipses are colored. Excerpt from the birds-eye-view grid at a distance of $30$ to $50$ meters from the ego vehicle.}
	\label{fig_attn_example}
\end{figure}

One approach towards joint object detection and tracking is the usage of transformer models \cite{vaswani_attention_2017}, which were able to achieve state-of-the-art results in some domains \cite{zhu2020deformable, meinhardt_trackformer:_2021}. Transformers are based on attention, i.e. the interaction between input tokens, which is why these models allow for a unique insight into their reasoning: One can visualize the attention matrices that are computed in each model forward pass and investigate which part of the input data was used to generate a certain output. In previous work, we developed a transformer based model for detection and tracking \cite{ruppel2022trans} in the context of autonomous driving that operates on (lidar) point clouds. An example of visualized attention weights from the tracking model are pictured in Figure~\ref{fig_attn_example}. In empirical observations, a more focused attention tends to lead to a more accurate detection. Therefore, we investigate whether the attention weight distribution can give insights into a detection uncertainty in this paper. An uncertainty indicator would be very valuable towards the ability to use transformer based methods in safety-critical use cases, such as autonomous driving.

The main contributions of this work are:
\begin{itemize}
	\item We propose a new metric, namely attention spread, to quantify the distribution of attention weights.
	\item We analyze whether attention spread is an indicator for uncertainty by comparing it to the observed detection accuracy in terms of Intersection-over-Union (IoU) with ground truth bounding boxes.
	\item We analyze the spatial and temporal dependencies of the observed attention weight matrices, both in the context of object detection as well as tracking.
\end{itemize}

\section{BACKGROUND}

\subsection{Attention in Transformers}
The transformer model was first introduced in \cite{vaswani_attention_2017} in the context of natural language processing, but has since been applied in many fields, such as computer vision \cite{carion_end--end_2020, dosovitskiy_image_2020}. The original model consists of an encoder and a decoder. In the encoder, a sequence of tokens is input, which can be encoded words, pixels, or grid cells, depending on the usage. These input tokens $\bm{x}_i$ (feature vectors), $i=~1,\dots, N$, can interact with one another through \textit{self-attention}. For this, each of them is transformed into three unique vectors via learnt mappings: a query $\bm{q}_i$, a key $\bm{k}_i$ and a value $\bm{v}_i$. Now, attention weights are computed by comparing each of the queries with each of the keys in terms of their dot product, obtaining scalar weights $w_{ij}=\bm{q}_i^\top\bm{k}_j$, which result in an \textit{attention weight matrix} of size $N\times N$. The output of the self-attention layer is obtained by a weighted sum of the values $\sum_{j=1}^{N}w_{ij}\bm{v}_j$, $i=1,\dots,N$, whereas computations are commonly performed in matrix form to increase efficiency and a softmax is applied to the weights to normalize their sum to $0$.

In the transformer decoder on the other hand, it is common to input two sequences: the encoder's output of length $N$, as well as a sequence of query vectors of length $M$. During \textit{self-attention} in the decoder, the queries attend to one another, while \textit{cross-attention} layers allow an interaction between the two sequences: In the attention module, the keys and values are now computed from one sequence, while the queries stem from the other. The attention weight matrices of size $N\times M$ during cross-attention in the transformer decoder are the focus of this paper.

\subsection{Transformers for Object Detection and Tracking}
The detection transformer (DETR) \cite{carion_end--end_2020} was one of the first models to utilize a transformer for object detection. It has been adapted and extended in multiple ways, for example for joint tracking and detection \cite{meinhardt_trackformer:_2021}. Our model from previous work \cite{ruppel2022trans}, which is the subject of the analysis in this paper, is based on the aforementioned approaches, but with a focus on applicability to large point clouds, such as those, which are commonly measured with automotive lidar sensors. The model is introduced in the following.

\begin{figure*}[!t]
	\includegraphics[width=\textwidth]{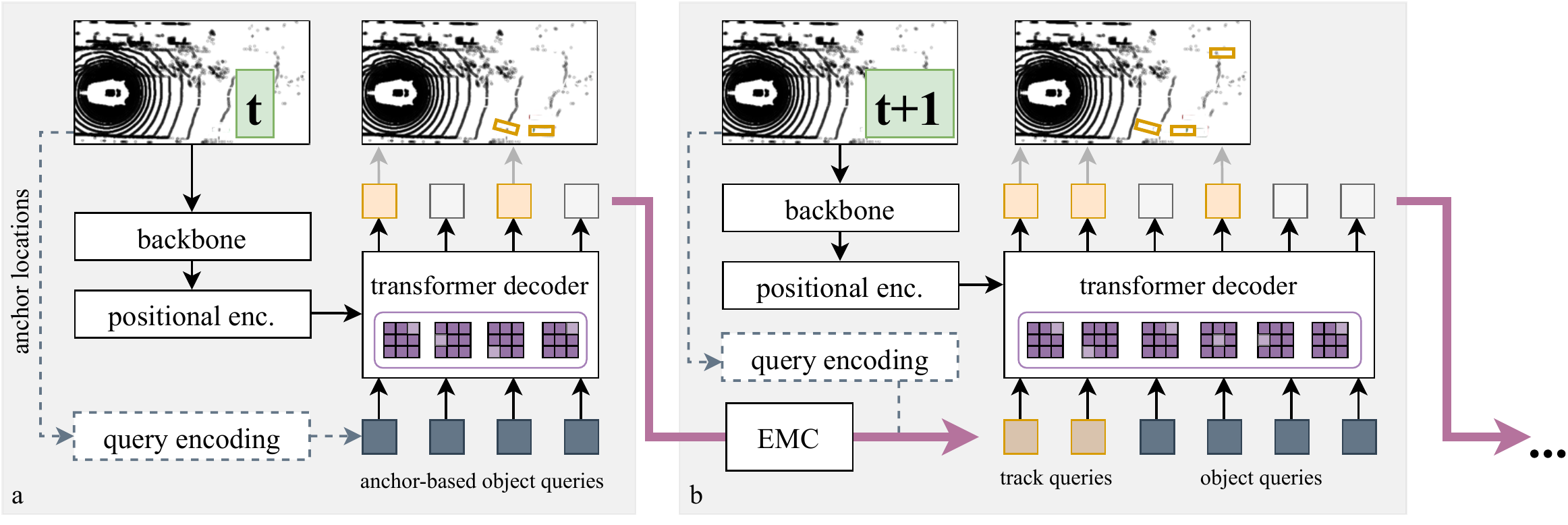}
	\caption{Detection and tracking model overview. The model design stems from previous work \cite{ruppel2022trans}. The focus of this paper lies with the analysis of the \textit{attention matrices} within the transformer decoder, which are illustrated in purple. a. Object detection model that can either be used as a standalone detector, or as track initializor in the first frame of a sequence. The input point cloud is processed through a backbone and a positional encoding is added. $M$ locations are sampled from the input point cloud and an object query is generated from each. In the transformer decoder, the queries can each aggregate information about one object, through cross-attention with the input feature map. b. Tracking model. Those feature vectors from the previous frame that belong to an object to be tracked are passed into the transformer decoder as track queries, in addition to the object queries. To compensate ego-motion between frames, they are transformed in feature space through an ego-motion compensation module (EMC).}
	\label{fig_model_overview}
\end{figure*}

An overview of the joint detection and tracking model is pictured in Figure~\ref{fig_model_overview}, with the detector on the left (a). This detector can either be used as a standalone model, i.e. without tracking, or it serves as a track initializor on the first frame of a sequence. The input point cloud is processed through a backbone, Pointpillars \cite{lang_pointpillars:_2019}. This could be replaced by any pretrained backbone that encodes the input into a grid (or sequence) of feature vectors. A positional encoding is added before the encoded input is passed to the transformer decoder as an unordered sequence. Note that the transformer \textit{encoder} is left out in this model, allowing for a comparably smaller GPU memory requirement. Therefore, only the backbone is available to provide context-encoding functionality. The second sequence that is passed to the decoder is denoted anchor-based object queries. $M$ queries are generated, which is more than the number of objects that are expected to appear in one frame. Each of them serves as a slot for a possible object and is able to access the input data through cross-attention, as well as interact with the other object queries during self-attention. Following \cite{misra_end--end_2021, ruppel2022det}, the query encoding is computed from prior locations, which are sampled from the input point cloud using farthest point sampling. This is meant to achieve an even spread of queries over the birds-eye-view grid, while not placing queries in areas where no lidar data is available (e.g. behind a large building). The object queries are transformed through $L$ decoder layers. To the resulting feature vectors, a regression and classification head is applied to obtain bounding box parameters of detected objects, whereas some are assigned to the 'no-object' class, since there are generally more queries than objects in the scene.

The focus of this paper are the cross-attention matrices in the decoder, of size $N\times M$. Since we operate on a square grid that is output by the backbone, we reshape the matrices to $\sqrt{N}\times\sqrt{N}\times M$, so that the first two dimensions correspond to indices on the birds-eye-view grid. For each object that is output by the model, one can obtain $L$ attention matrices of shape $\sqrt{N}\times\sqrt{N}$ (one per decoder layer), giving an insight, which input data the respective query accessed to detect this object.

In Figure~\ref{fig_model_overview}~(b), the object tracker \cite{ruppel2022trans} is pictured. In addition to the object queries, track queries are passed to the decoder, which serve as slots for continued tracks. These stem from the model's output at the previous timestep and contain information about an object in feature space. They are transformed through an ego-motion compensation module (EMC) to correct the ego-motion between frames. For these track queries, the aforementioned attention weight matrices can be obtained as well.

\section{PROPOSED METHOD}
\subsection{Modelling the attention distribution}
The cross-attention weights between one object or track query and the input birds-eye-view grid in a given decoder layer are reshaped to a matrix of shape $\sqrt{N}\times \sqrt{N}$, as introduced above. It contains attention weights $w_{pq}$, $1\leq p,q \leq \sqrt{N}$, where $p$ and $q$ correspond to indices on the birds-eye-view grid. The largest $K$ weights are selected, i.e. $p,q\in S^K$. By definition of the birds-eye-view grid, each horizontal index $q$ is associated with a location in $x$-direction $x_q$, and each index $p$ with one in $y$-direction, $y_p$. We define the mean of top-$K$ attention as
\begin{equation}
\bm{\mu}_{K} = \frac{1}{W}\sum_{p,q\in S^K} w_{p,q} \begin{pmatrix}x_q \\ y_p\end{pmatrix},
\end{equation}
where $W=\sum_{p,q\in S^K}w_{p,q}$.
We propose to quantify attention covariance as follows:
\begin{equation}
\bm{C}_K=\frac{1}{W}\sum_{p,q\in S^K} w_{p,q}\left[\begin{pmatrix}x_q \\ y_p\end{pmatrix}-\bm{\mu}_{K}\right]\left[\begin{pmatrix}x_q \\ y_p\end{pmatrix}-\bm{\mu}_{K}\right]^\top.
\end{equation}
This covariance matrix can be reduced to a single value, namely the proposed attention spread (AS)
\begin{equation}
\textrm{AS}= \det{\bm{C}_K}.
\end{equation}
It is difficult to quantify 'ground truth' uncertainty of a deep learning model. In this work, we quantify detection accuracy in terms of intersection-over-union (IoU) between estimated boxes and ground truth. IoU has been shown to correlate with epistemic uncertainty \cite{feng2018uncertainty}. Aleatoric uncertainty, on the other hand, arises with low visibility and object distance \cite{wirges}, among other causes.
\subsection{Attention shift and focus}
In the detector, an object query passes through $L$ decoder layers. Before the first layer, it is assigned an anchor location, from which its encoding is computed and that serves as a prior location for object detection. One would expect that the attention between the query and the input data in the first layer is quite broad, since it might move away to find a nearby object, while still located around the prior location. As the object query continues to focus on one object through the following layers, the resulting attention is expected to be more focused around that object. We propose to test these assumptions by collecting the aforementioned attention distribution parameters for each decoder layer.

Another related question of interest is how the attention focus changes during the lifetime of a track.
\section{RESULTS}
For this analysis, we set $M=100$, $\sqrt{N}=128$, $L=~6$, $K=~100$, and use the pretrained detection and tracking model as introduced in previous work \cite{ruppel2022trans}. The experiments are performed on the nuScenes {\tt val} dataset and only the class 'car' is considered. The model is either used in detection mode, i.e. without passing temporal information to the following frame, or in tracking mode, by passing track queries to the model in addition to the object queries. Only bounding boxes are considered that the model itself classified as relatively confident, with a detection score above $0.8$, since it is common in this model that many queries are idle.
\subsection{Detection and tracking accuracy}
\begin{figure}[t]
	\vspace{10pt}
\begin{tikzpicture}

\definecolor{color1}{rgb}{0.3803921568627451, 0.6627450980392157, 0.3607843137254902}
\definecolor{color0}{rgb}{1.0, 0.5725490196078431, 0.1411764705882353}

\begin{axis}[
height=151.86214363138544,
tick align=outside,
tick pos=left,
width=235.71811,
x grid style={white!69.0196078431373!black},
xmin=0.00999999959021806, xmax=0.890000024996698,
xlabel={IoU},
ylabel={Attention Spread},
xtick style={color=black},
xtick={0,0.1,0.2,0.3,0.4,0.5,0.6,0.7,0.8,0.9},
xticklabels={0.0,0.1,0.2,0.3,0.4,0.5,0.6,0.7,0.8,0.9},
y grid style={white!69.0196078431373!black},
ymin=3, ymax=537.480094265938,
ymode=log,
ytick style={color=black}
]
\path [draw=color0, semithick]
(axis cs:0.0500000007450581,70.9432106018066)
--(axis cs:0.0500000007450581,438.134780883789);

\path [draw=color0, semithick]
(axis cs:0.150000005960464,60.9441070556641)
--(axis cs:0.150000005960464,401.912178039551);

\path [draw=color0, semithick]
(axis cs:0.25,53.5082206726074)
--(axis cs:0.25,343.74609375);

\path [draw=color0, semithick]
(axis cs:0.350000023841858,49.8880128860474)
--(axis cs:0.350000023841858,296.952857971191);

\path [draw=color0, semithick]
(axis cs:0.450000017881393,38.7504253387451)
--(axis cs:0.450000017881393,235.086570739746);

\path [draw=color0, semithick]
(axis cs:0.550000011920929,24.7958936691284)
--(axis cs:0.550000011920929,169.361778259277);

\path [draw=color0, semithick]
(axis cs:0.650000035762787,12.6368269920349)
--(axis cs:0.650000035762787,101.787349700928);

\path [draw=color0, semithick]
(axis cs:0.75,6.43029975891113)
--(axis cs:0.75,58.5919513702393);

\path [draw=color0, semithick]
(axis cs:0.850000023841858,4.30279064178467)
--(axis cs:0.850000023841858,42.45485496521);

\addplot [semithick, color0, mark=-, mark size=3, mark options={solid}, only marks]
table {%
	0.0500000007450581 70.9432106018066
	0.150000005960464 60.9441070556641
	0.25 53.5082206726074
	0.350000023841858 49.8880128860474
	0.450000017881393 38.7504253387451
	0.550000011920929 24.7958936691284
	0.650000035762787 12.6368269920349
	0.75 6.43029975891113
	0.850000023841858 4.30279064178467
};
\addplot [semithick, color0, mark=-, mark size=3, mark options={solid}, only marks]
table {%
	0.0500000007450581 438.134780883789
	0.150000005960464 401.912178039551
	0.25 343.74609375
	0.350000023841858 296.952857971191
	0.450000017881393 235.086570739746
	0.550000011920929 169.361778259277
	0.650000035762787 101.787349700928
	0.75 58.5919513702393
	0.850000023841858 42.45485496521
};
\addplot [semithick, color0]
table {%
	0.0500000007450581 121.050827026367
	0.150000005960464 94.6852035522461
	0.25 84.3253936767578
	0.350000023841858 79.9751815795898
	0.450000017881393 66.3462600708008
	0.550000011920929 46.1697120666504
	0.650000035762787 29.2016181945801
	0.75 19.8154029846191
	0.850000023841858 16.2230186462402
};
\addplot [name path=upper0,draw=none] table{%
	0.0500000007450581 438.134780883789
	0.150000005960464 401.912178039551
	0.25 343.74609375
	0.350000023841858 296.952857971191
	0.450000017881393 235.086570739746
	0.550000011920929 169.361778259277
	0.650000035762787 101.787349700928
	0.75 58.5919513702393
	0.850000023841858 42.45485496521
};
\addplot [name path=lower0,draw=none] table{%
	0.0500000007450581 70.9432106018066
	0.150000005960464 60.9441070556641
	0.25 53.5082206726074
	0.350000023841858 49.8880128860474
	0.450000017881393 38.7504253387451
	0.550000011920929 24.7958936691284
	0.650000035762787 12.6368269920349
	0.75 6.43029975891113
	0.850000023841858 4.30279064178467
};
\addplot [fill=color0!10] fill between[of=upper0 and lower0];
\end{axis}
\vspace{10pt}
\end{tikzpicture}
\begin{tikzpicture}

\definecolor{color0}{rgb}{0.3803921568627451, 0.6627450980392157, 0.3607843137254902}
\definecolor{color1}{rgb}{1.0, 0.5725490196078431, 0.1411764705882353}

\begin{axis}[
height=151.86214363138544,
tick align=outside,
tick pos=left,
width=235.71811,
x grid style={white!69.0196078431373!black},
xmin=0.00999999959021806, xmax=0.890000024996698,
xlabel={IoU},
ylabel={Attention spread},
xtick style={color=black},
xtick={0,0.1,0.2,0.3,0.4,0.5,0.6,0.7,0.8,0.9},
xticklabels={0.0,0.1,0.2,0.3,0.4,0.5,0.6,0.7,0.8,0.9},
y grid style={white!69.0196078431373!black},
ymin=3, ymax=537.480094265938,
ymode=log,
ytick style={color=black}
]
\path [draw=color0, semithick]
(axis cs:0.0500000007450581,90.8004169464111)
--(axis cs:0.0500000007450581,512.092811584473);

\path [draw=color0, semithick]
(axis cs:0.150000005960464,65.9673461914062)
--(axis cs:0.150000005960464,413.120475769043);

\path [draw=color0, semithick]
(axis cs:0.25,55.2956008911133)
--(axis cs:0.25,344.91918182373);

\path [draw=color0, semithick]
(axis cs:0.350000023841858,46.191349029541)
--(axis cs:0.350000023841858,296.456867218018);

\path [draw=color0, semithick]
(axis cs:0.450000017881393,41.3171815872192)
--(axis cs:0.450000017881393,252.070930480957);

\path [draw=color0, semithick]
(axis cs:0.550000011920929,25.5156707763672)
--(axis cs:0.550000011920929,174.41872215271);

\path [draw=color0, semithick]
(axis cs:0.650000035762787,13.5695085525513)
--(axis cs:0.650000035762787,107.186096191406);

\path [draw=color0, semithick]
(axis cs:0.75,6.86533379554749)
--(axis cs:0.75,62.6639318466187);

\path [draw=color0, semithick]
(axis cs:0.850000023841858,4.34715795516968)
--(axis cs:0.850000023841858,42.1914620399475);

\addplot [semithick, color0, mark=-, mark size=3, mark options={solid}, only marks]
table {%
	0.0500000007450581 90.8004169464111
	0.150000005960464 65.9673461914062
	0.25 55.2956008911133
	0.350000023841858 46.191349029541
	0.450000017881393 41.3171815872192
	0.550000011920929 25.5156707763672
	0.650000035762787 13.5695085525513
	0.75 6.86533379554749
	0.850000023841858 4.34715795516968
};
\addplot [semithick, color0, mark=-, mark size=3, mark options={solid}, only marks]
table {%
	0.0500000007450581 512.092811584473
	0.150000005960464 413.120475769043
	0.25 344.91918182373
	0.350000023841858 296.456867218018
	0.450000017881393 252.070930480957
	0.550000011920929 174.41872215271
	0.650000035762787 107.186096191406
	0.75 62.6639318466187
	0.850000023841858 42.1914620399475
};
\addplot [semithick, color0]
table {%
	0.0500000007450581 141.334655761719
	0.150000005960464 103.826347351074
	0.25 85.9954452514648
	0.350000023841858 76.2963409423828
	0.450000017881393 68.5205078125
	0.550000011920929 47.2218399047852
	0.650000035762787 30.3506965637207
	0.75 20.6654167175293
	0.850000023841858 16.3191566467285
};
\addplot [name path=upper0,draw=none] table{%
	0.0500000007450581 512.092811584473
	0.150000005960464 413.120475769043
	0.25 344.91918182373
	0.350000023841858 296.456867218018
	0.450000017881393 252.070930480957
	0.550000011920929 174.41872215271
	0.650000035762787 107.186096191406
	0.75 62.6639318466187
	0.850000023841858 42.1914620399475
};
\addplot [name path=lower0,draw=none] table{%
	0.0500000007450581 90.8004169464111
	0.150000005960464 65.9673461914062
	0.25 55.2956008911133
	0.350000023841858 46.191349029541
	0.450000017881393 41.3171815872192
	0.550000011920929 25.5156707763672
	0.650000035762787 13.5695085525513
	0.75 6.86533379554749
	0.850000023841858 4.34715795516968
};
\addplot [fill=color0!10] fill between[of=upper0 and lower0];
\end{axis}
\vspace{10pt}
\end{tikzpicture}
	\caption{Median attention spread in terms of intersection-over-union between the estimated bounding box and the closest ground truth box, with $25$th and $75$th percentiles as error bars. IoU values were grouped into bins of width $0.1$, while those smaller than $0$ were excluded. \textit{Top:} Object detection \textit{Bottom:} Joint detection and tracking.}
	\label{fig_AS_IOU}
\end{figure}
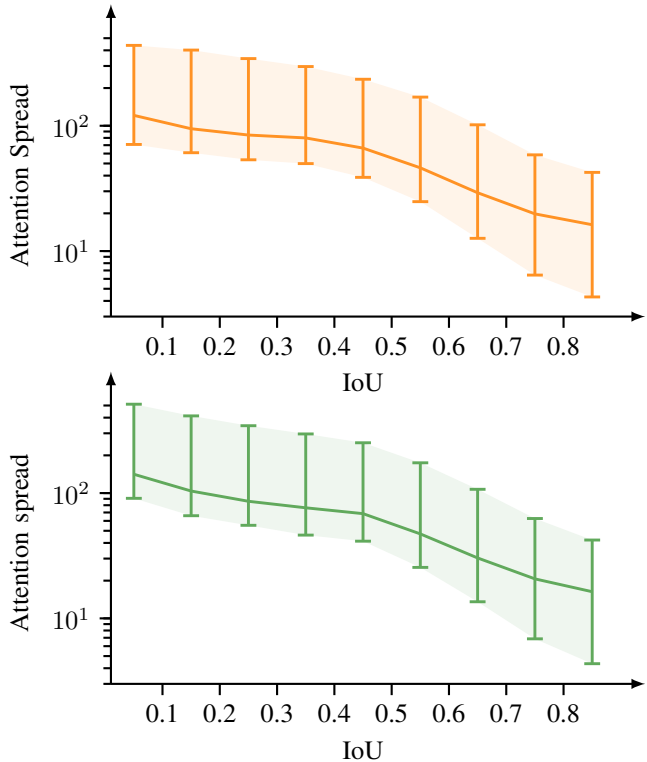

In Figure~\ref{fig_AS_IOU}, attention spread is plotted in terms of intersection-over-union (IoU). For this, the IoU between the estimated bounding boxes and their closest ground truth object was computed. A large IoU corresponds to a more accurately detected object. IoU values of $0$ (no overlap) were removed. The IoU values were grouped into 10 bins of width $0.1$, and the corresponding attention spread obtained from the last decoder layer. The median attention spread per bin was plotted, as well as $25$th and $75$th percentiles as error bars. The plots are very similar for object detection (top) and joint detection and tracking (bottom). The observed median attention spread sinks with growing IoU. This means that low attention spread can be an indicator for high IoU and vice-versa. Therefore, attention spread may also indicate epistemic uncertainty \cite{feng2018uncertainty}, however the ranges between $25$th and $75$th percentiles are quite large.
\subsection{Object distance to the ego vehicle}
\begin{figure}[t]
\begin{tikzpicture}

\begin{axis}[
colorbar,
colorbar style={align=center, ylabel={Average attention spread}, ymode=log, axis y line=right, y axis line style={-, shorten >=0cm,thin}, x axis line style={opacity=0}, tick style=thin},
colormap/viridis,
axis line style={opacity=0},
height=185,
point meta max=69878.4375809,
point meta min=64.2002339655,
tick align=outside,
width=185,
x grid style={white!69.0196078431373!black},
xlabel={Distance in meters (x)},
xmin=-51.2, xmax=51.2,
xtick pos=both,
xtick style={color=black},
y grid style={white!69.0196078431373!black},
ylabel={Distance in meters (y)},
ymin=-51.2, ymax=51.2,
ytick pos=left,
ytick style={color=black}
]
\addplot graphics [includegraphics cmd=\pgfimage,xmin=-51.2, xmax=51.2, ymin=-51.2, ymax=51.2] {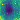};
\end{axis}
\end{tikzpicture}
	\vspace{-12pt}
	\caption{Attention spread in terms of the locations of estimated bounding boxes in birds-eye-view. Locations were grouped into $20\times 20$ bins and their respective attention spreads averaged.}
	\label{fig_2d_cov}
\end{figure}

To see whether attention spread changes with the distance of estimated bounding boxes to the ego vehicle, the birds-eye-view is divided into $20\times 20$ bins. For each bin, estimated bounding boxes located there, as well as their respective attention spread in the last layer of the detector, are aggregated. Their mean per bin over all frames in the nuScenes {\tt val} dataset is pictured in Figure~\ref{fig_2d_cov}. It is observable that mean attention spread increases with distance to the ego vehicle. In this regard, it behaves similarly to aleatoric uncertainty \cite{wirges}.
\subsection{Attention through multiple decoder layers}
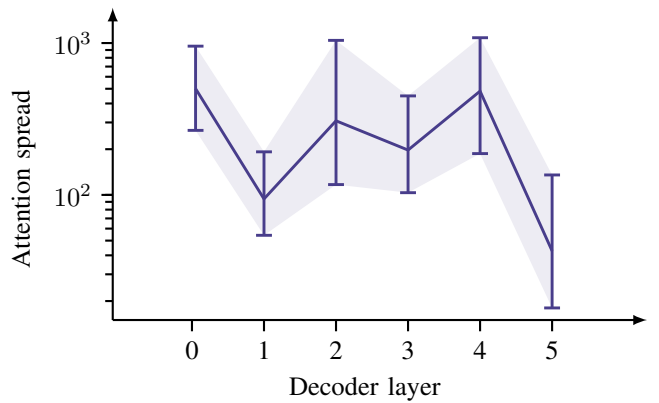
\begin{figure}[t]
	\vspace{10pt}
\begin{tikzpicture}

\definecolor{color0}{rgb}{0.2823529411764706, 0.23921568627450981, 0.5450980392156862}
\definecolor{color1}{rgb}{1.0, 0.5725490196078431, 0.1411764705882353}

\begin{axis}[
height=151.86214363138544,
tick align=outside,
tick pos=left,
width=235.71811,
x grid style={white!69.0196078431373!black},
xmin=-1.0999999959021806, xmax=5.90000024996698,
xlabel={Decoder layer},
ylabel={Attention spread},
xtick style={color=black},
xtick={0,1,2,3,4,5},
xticklabels={0,1,2,3,4,5},
y grid style={white!69.0196078431373!black},
ymin=15,
ymode=log,
ytick style={color=black}
]
\path [draw=color0, semithick]
(axis cs:0.0500000007450581,266.07301330566406)
--(axis cs:0.0500000007450581,954.1417541503906);

\path [draw=color0, semithick]
(axis cs:4,187.1455535888672)
--(axis cs:4,1084.18310546875);

\path [draw=color0, semithick]
(axis cs:2,116.94940567016602)
--(axis cs:2,1041.6171264648438);

\path [draw=color0, semithick]
(axis cs:3,103.42736434936523)
--(axis cs:3,448.7355194091797);

\path [draw=color0, semithick]
(axis cs:1,54.201114654541016)
--(axis cs:1,192.1822052001953);

\path [draw=color0, semithick]
(axis cs:5,18.0173282623291)
--(axis cs:5,135.3311309814453);

\addplot [semithick, color0, mark=-, mark size=3, mark options={solid}, only marks]
table {%
	0.0500000007450581 266.07301330566406
	1 54.201114654541016
	2 116.94940567016602
	3 103.42736434936523
	4 187.1455535888672
	5 18.0173282623291
};
\addplot [semithick, color0, mark=-, mark size=3, mark options={solid}, only marks]
table {%
	0.0500000007450581 954.1417541503906
	1 192.1822052001953
	2 1041.6171264648438
	3 448.7355194091797
	4 1084.18310546875
	5 135.3311309814453
};
\addplot [semithick, color0]
table {%
	0.0500000007450581 501.1051
	1 94.07015
	2 307.63257
	3 197.37721
	4 482.34073
	5 42.76048
};
\addplot [name path=upper0,draw=none] table{%
	0.0500000007450581 954.1417541503906
	1 192.1822052001953
	2 1041.6171264648438
	3 448.7355194091797
	4 1084.18310546875
	5 135.3311309814453
};
\addplot [name path=lower0,draw=none] table{%
	0.0500000007450581 266.07301330566406
	1 54.201114654541016
	2 116.94940567016602
	3 103.42736434936523
	4 187.1455535888672
	5 18.0173282623291
};
\addplot [fill=color0!10] fill between[of=upper0 and lower0];
\end{axis}
\vspace{10pt}
\end{tikzpicture}
	\caption{Median attention spread per layer in the transformer decoder. The error bars denote $25$th and $75$th percentiles.}
	\label{fig_spread_over_layer}
\end{figure}

To investigate how attention spread changes through the decoder layers, its values observed during object detection are collected per layer for each observed object. Median attention spread per layer is pictured in Figure~\ref{fig_spread_over_layer}, as well as its $25$th and $75$th percentiles. The mean attention spread in the last layer is smaller than in the first, which is in line with the expectations. However, the value is quite small in the second layer and rises again in the following, which poses the question whether there exists a specific reason for this early focusing and later broadening of the cross-attention.
\subsection{Attention during a track's lifetime}
Observing many tracked objects, we find that the attention spread of a tracked object changes over time. One exemplary visualization of the initialization phase of a track is pictured in Figure~\ref{fig_track_attent}. When the lime green track is initialized at a distance of ca. 50 meters, it barely overlaps with the field of view. Its attention spread is quite large (displayed in terms of an ellipse that is derived from the covariance matrix). As it moves closer over time, its attention spread decreases.

In Figure~\ref{fig_track_spread} (left), the median attention spread in terms of elapsed time since track initialization is pictured. For this, the first $3.5$ seconds of all tracks, as output by the model on the nuScenes {\tt val} dataset, were considered if the total length of the respective track was at least $7$ seconds. In Figure~\ref{fig_track_spread} (right), the median attention spread in terms of the remaining time until track finalization is pictured. In both figures, error bars denote the $25$th and $75$th percentiles.

\addtolength{\textheight}{-4.5cm}   
We observe that both in the track initialization phase as well as in the finalization phase a change in median attention is visible. Multiple factors may influence this behavior: Firstly, tracks are often initialized and finalized at a large distance to the ego vehicle, when an object first enters and last exists the field of view, respectively. In that regard, the observations are similar to those in Figure~\ref{fig_2d_cov}. Second, tracks may increase in confidence after a certain initialization time, since the model can utilize past observations, encoded in the track queries, for its reasoning.
\begin{figure}
	\centering
	\begin{tabular}{ c @{\hspace{10pt}} c @{\hspace{10pt}} c }
		\includegraphics[width=.3\columnwidth]{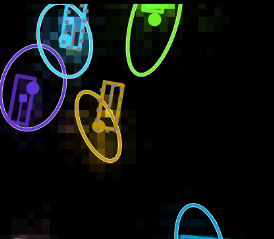} &
		\includegraphics[width=.3\columnwidth]{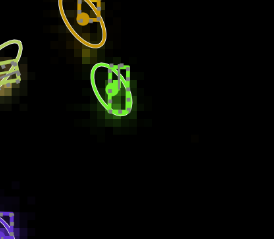} &
		\includegraphics[width=.3\columnwidth]{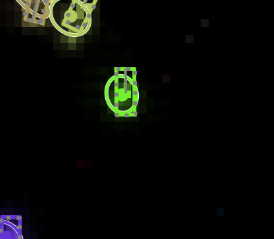}\\
		\small 50 meters &
		\small 30 meters &
		\small 10 meters
	\end{tabular}
	
	\medskip
	
	\caption{Exemplary attention during the lifetime of one track (light green). The track is initialized at a distance of ca. $50$ meters and then moves closer towards the ego vehicle.}
	\label{fig_track_attent}
\end{figure}
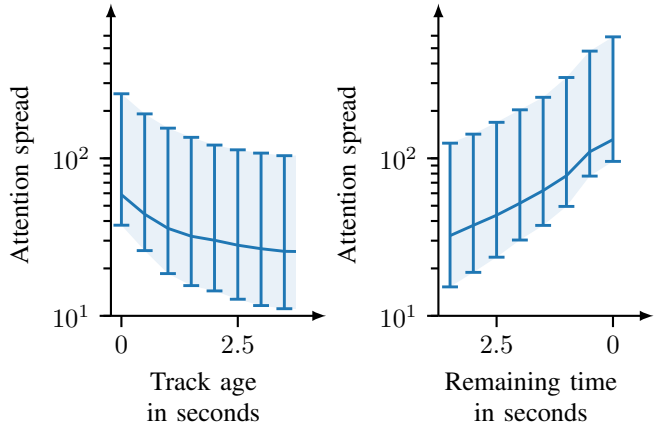
\begin{figure}[t]
	\vspace{10pt}
\begin{tikzpicture}

\definecolor{color0}{rgb}{0.12156862745098,0.466666666666667,0.705882352941177}

\begin{axis}[
height=151.86214363138544,
tick align=outside,
tick pos=left,
width=115,
x grid style={white!69.0196078431373!black},
xmin=-0.45, xmax=7.5,
xtick style={color=black},
y grid style={white!69.0196078431373!black},
ymin=10, ymax=619.597747778893,
ytick={1e1,1e2, 1e3},
ylabel={Attention spread},
align=center,
xlabel={Track age\\ in seconds},
scaled x ticks={real:2},
xtick scale label code/.code={},
ymode=log,
ytick style={color=black},
]
\path [draw=color0, semithick]
(axis cs:0,37.6250581741333)
--(axis cs:0,256.879245758057);

\path [draw=color0, semithick]
(axis cs:1,25.9533352851868)
--(axis cs:1,191.472328186035);

\path [draw=color0, semithick]
(axis cs:2,18.5453748703003)
--(axis cs:2,155.418348312378);

\path [draw=color0, semithick]
(axis cs:3,15.5591969490051)
--(axis cs:3,135.901071548462);

\path [draw=color0, semithick]
(axis cs:4,14.3977401256561)
--(axis cs:4,121.36810874939);

\path [draw=color0, semithick]
(axis cs:5,12.7488734722137)
--(axis cs:5,113.215627670288);

\path [draw=color0, semithick]
(axis cs:6,11.6232378482819)
--(axis cs:6,107.928728103638);

\path [draw=color0, semithick]
(axis cs:7,11.0989985466003)
--(axis cs:7,103.938653945923);

\path [draw=color0, semithick]
(axis cs:8,11.0127696990967)
--(axis cs:8,104.324741363525);

\path [draw=color0, semithick]
(axis cs:9,11.5648920536041)
--(axis cs:9,107.340913772583);

\path [draw=color0, semithick]
(axis cs:10,11.9008374214172)
--(axis cs:10,114.132675170898);

\path [draw=color0, semithick]
(axis cs:11,14.071426153183)
--(axis cs:11,121.763483047485);

\path [draw=color0, semithick]
(axis cs:12,15.2912001609802)
--(axis cs:12,133.930896759033);

\path [draw=color0, semithick]
(axis cs:13,18.9306454658508)
--(axis cs:13,155.243856430054);

\path [draw=color0, semithick]
(axis cs:14,21.710081577301)
--(axis cs:14,183.835567474365);

\path [draw=color0, semithick]
(axis cs:15,26.919641494751)
--(axis cs:15,221.03438949585);

\path [draw=color0, semithick]
(axis cs:16,27.2940921783447)
--(axis cs:16,218.016418457031);

\path [draw=color0, semithick]
(axis cs:17,28.2429947853088)
--(axis cs:17,220.044715881348);

\path [draw=color0, semithick]
(axis cs:18,26.3938698768616)
--(axis cs:18,220.956077575684);

\path [draw=color0, semithick]
(axis cs:19,28.1997604370117)
--(axis cs:19,221.54317855835);

\path [draw=color0, semithick]
(axis cs:20,29.7406902313232)
--(axis cs:20,225.525279998779);

\path [draw=color0, semithick]
(axis cs:21,30.6817226409912)
--(axis cs:21,230.070346832275);

\path [draw=color0, semithick]
(axis cs:22,25.3371734619141)
--(axis cs:22,213.922431945801);

\path [draw=color0, semithick]
(axis cs:23,24.1157040596008)
--(axis cs:23,200.321170806885);

\path [draw=color0, semithick]
(axis cs:24,23.5696630477905)
--(axis cs:24,187.576179504395);

\path [draw=color0, semithick]
(axis cs:25,24.5937490463257)
--(axis cs:25,225.014377593994);

\path [draw=color0, semithick]
(axis cs:26,26.3718333244324)
--(axis cs:26,213.710731506348);

\path [draw=color0, semithick]
(axis cs:27,22.7520313262939)
--(axis cs:27,176.121139526367);

\path [draw=color0, semithick]
(axis cs:28,18.5318832397461)
--(axis cs:28,182.438426971436);

\path [draw=color0, semithick]
(axis cs:29,13.6331491470337)
--(axis cs:29,148.564577102661);

\path [draw=color0, semithick]
(axis cs:30,17.889922618866)
--(axis cs:30,139.027460098267);

\path [draw=color0, semithick]
(axis cs:31,14.7497441768646)
--(axis cs:31,118.954608917236);

\path [draw=color0, semithick]
(axis cs:32,15.5837469100952)
--(axis cs:32,153.274448394775);

\path [draw=color0, semithick]
(axis cs:33,13.2620630264282)
--(axis cs:33,136.92121887207);

\path [draw=color0, semithick]
(axis cs:34,11.2035417556763)
--(axis cs:34,140.847946166992);

\path [draw=color0, semithick]
(axis cs:35,12.746621131897)
--(axis cs:35,136.361822128296);

\path [draw=color0, semithick]
(axis cs:36,11.0700778961182)
--(axis cs:36,136.357130050659);

\path [draw=color0, semithick]
(axis cs:37,12.0639040470123)
--(axis cs:37,139.791702270508);

\path [draw=color0, semithick]
(axis cs:38,10.2564690113068)
--(axis cs:38,111.135135650635);

\path [draw=color0, semithick]
(axis cs:39,9.008704662323)
--(axis cs:39,96.4599113464355);

\addplot [semithick, color0, mark=-, mark size=3, mark options={solid}, only marks]
table {%
0 37.6250581741333
1 25.9533352851868
2 18.5453748703003
3 15.5591969490051
4 14.3977401256561
5 12.7488734722137
6 11.6232378482819
7 11.0989985466003
8 11.0127696990967
9 11.5648920536041
10 11.9008374214172
11 14.071426153183
12 15.2912001609802
13 18.9306454658508
14 21.710081577301
15 26.919641494751
16 27.2940921783447
17 28.2429947853088
18 26.3938698768616
19 28.1997604370117
20 29.7406902313232
21 30.6817226409912
22 25.3371734619141
23 24.1157040596008
24 23.5696630477905
25 24.5937490463257
26 26.3718333244324
27 22.7520313262939
28 18.5318832397461
29 13.6331491470337
30 17.889922618866
31 14.7497441768646
32 15.5837469100952
33 13.2620630264282
34 11.2035417556763
35 12.746621131897
36 11.0700778961182
37 12.0639040470123
38 10.2564690113068
39 9.008704662323
};
\addplot [semithick, color0, mark=-, mark size=3, mark options={solid}, only marks]
table {%
0 256.879245758057
1 191.472328186035
2 155.418348312378
3 135.901071548462
4 121.36810874939
5 113.215627670288
6 107.928728103638
7 103.938653945923
8 104.324741363525
9 107.340913772583
10 114.132675170898
11 121.763483047485
12 133.930896759033
13 155.243856430054
14 183.835567474365
15 221.03438949585
16 218.016418457031
17 220.044715881348
18 220.956077575684
19 221.54317855835
20 225.525279998779
21 230.070346832275
22 213.922431945801
23 200.321170806885
24 187.576179504395
25 225.014377593994
26 213.710731506348
27 176.121139526367
28 182.438426971436
29 148.564577102661
30 139.027460098267
31 118.954608917236
32 153.274448394775
33 136.92121887207
34 140.847946166992
35 136.361822128296
36 136.357130050659
37 139.791702270508
38 111.135135650635
39 96.4599113464355
};
\addplot [semithick, color0]
table {%
0 58.8044815063477
1 44.1706428527832
2 35.9213562011719
3 32.0216979980469
4 30.1864204406738
5 28.0841102600098
6 26.6770286560059
7 25.675874710083
8 25.5513763427734
9 26.282958984375
10 26.7699413299561
11 29.6150226593018
12 32.3285827636719
13 36.4808883666992
14 40.4689559936523
15 46.3212356567383
16 46.58251953125
17 47.3008651733398
18 45.4139862060547
19 46.7853469848633
20 49.013988494873
21 49.0651092529297
22 44.0095748901367
23 41.8960037231445
24 41.7168655395508
25 43.0303611755371
26 43.9225540161133
27 38.9740447998047
28 35.1348495483398
29 30.6218509674072
30 33.4381294250488
31 30.7304267883301
32 32.016357421875
33 28.4360504150391
34 26.3786582946777
35 28.1391506195068
36 26.465856552124
37 27.3940086364746
38 25.5564708709717
39 25.225887298584
};
\addplot [name path=upper0,draw=none] table{%
	0 256.879245758057
	1 191.472328186035
	2 155.418348312378
	3 135.901071548462
	4 121.36810874939
	5 113.215627670288
	6 107.928728103638
	7 103.938653945923
	8 104.324741363525
};
\addplot [name path=lower0,draw=none] table{%
	0 37.6250581741333
	1 25.9533352851868
	2 18.5453748703003
	3 15.5591969490051
	4 14.3977401256561
	5 12.7488734722137
	6 11.6232378482819
	7 11.0989985466003
	8 11.0127696990967
};
\addplot [fill=color0!10] fill between[of=upper0 and lower0];
\end{axis}

\end{tikzpicture}\hspace{5pt}
\begin{tikzpicture}

\definecolor{color0}{rgb}{0.12156862745098,0.466666666666667,0.705882352941177}

\begin{axis}[
height=151.86214363138544,
tick align=outside,
tick pos=left,
width=115,
x grid style={white!69.0196078431373!black},
xmin=-7.5, xmax=0.45,
xtick style={color=black},
ylabel={Attention spread},
align=center,
xlabel={Remaining time\\ in seconds},
scaled x ticks={real:-2},
xtick scale label code/.code={},
y grid style={white!69.0196078431373!black},
ymin=10, ymax=619.597747778893,
ytick style={color=black},
ytick={1e1,1e2, 1e3},
ymode=log
]
\path [draw=color0, semithick]
(axis cs:-39,5.22390031814575)
--(axis cs:-39,74.4397230148315);

\path [draw=color0, semithick]
(axis cs:-38,6.25572681427002)
--(axis cs:-38,65.9752550125122);

\path [draw=color0, semithick]
(axis cs:-37,6.56444478034973)
--(axis cs:-37,80.6355037689209);

\path [draw=color0, semithick]
(axis cs:-36,6.09748029708862)
--(axis cs:-36,73.5315656661987);

\path [draw=color0, semithick]
(axis cs:-35,6.96366310119629)
--(axis cs:-35,76.6602458953857);

\path [draw=color0, semithick]
(axis cs:-34,9.30453300476074)
--(axis cs:-34,84.8390159606934);

\path [draw=color0, semithick]
(axis cs:-33,8.21129703521729)
--(axis cs:-33,98.0365562438965);

\path [draw=color0, semithick]
(axis cs:-32,8.64026260375977)
--(axis cs:-32,91.4644451141357);

\path [draw=color0, semithick]
(axis cs:-31,7.84352970123291)
--(axis cs:-31,92.6722507476807);

\path [draw=color0, semithick]
(axis cs:-30,8.75040245056152)
--(axis cs:-30,87.4636917114258);

\path [draw=color0, semithick]
(axis cs:-29,8.84356427192688)
--(axis cs:-29,95.9056987762451);

\path [draw=color0, semithick]
(axis cs:-28,10.8798811435699)
--(axis cs:-28,110.316257476807);

\path [draw=color0, semithick]
(axis cs:-27,11.1100029945374)
--(axis cs:-27,105.531917572021);

\path [draw=color0, semithick]
(axis cs:-26,10.1690502166748)
--(axis cs:-26,107.004907608032);

\path [draw=color0, semithick]
(axis cs:-25,10.3997278213501)
--(axis cs:-25,106.592775344849);

\path [draw=color0, semithick]
(axis cs:-24,12.4184679985046)
--(axis cs:-24,110.657705307007);

\path [draw=color0, semithick]
(axis cs:-23,11.3108296394348)
--(axis cs:-23,107.920598983765);

\path [draw=color0, semithick]
(axis cs:-22,11.7723472118378)
--(axis cs:-22,111.723009109497);

\path [draw=color0, semithick]
(axis cs:-21,11.1935997009277)
--(axis cs:-21,107.256378173828);

\path [draw=color0, semithick]
(axis cs:-20,12.8980417251587)
--(axis cs:-20,112.990243911743);

\path [draw=color0, semithick]
(axis cs:-19,13.4059295654297)
--(axis cs:-19,113.3626537323);

\path [draw=color0, semithick]
(axis cs:-18,11.8703980445862)
--(axis cs:-18,107.675174713135);

\path [draw=color0, semithick]
(axis cs:-17,12.3548753261566)
--(axis cs:-17,113.856777191162);

\path [draw=color0, semithick]
(axis cs:-16,12.2391397953033)
--(axis cs:-16,116.550710678101);

\path [draw=color0, semithick]
(axis cs:-15,14.7252283096313)
--(axis cs:-15,124.263000488281);

\path [draw=color0, semithick]
(axis cs:-14,13.0979306697845)
--(axis cs:-14,119.959365844727);

\path [draw=color0, semithick]
(axis cs:-13,11.6684587001801)
--(axis cs:-13,113.417181015015);

\path [draw=color0, semithick]
(axis cs:-12,11.5907690525055)
--(axis cs:-12,103.946506500244);

\path [draw=color0, semithick]
(axis cs:-11,12.3287258148193)
--(axis cs:-11,102.013608932495);

\path [draw=color0, semithick]
(axis cs:-10,10.5150499343872)
--(axis cs:-10,98.5654792785645);

\path [draw=color0, semithick]
(axis cs:-7,15.2682423591614)
--(axis cs:-7,124.790456771851);

\path [draw=color0, semithick]
(axis cs:-6,18.8793020248413)
--(axis cs:-6,142.401453018188);

\path [draw=color0, semithick]
(axis cs:-5,23.5670943260193)
--(axis cs:-5,169.25256729126);

\path [draw=color0, semithick]
(axis cs:-4,30.2904300689697)
--(axis cs:-4,203.178112030029);

\path [draw=color0, semithick]
(axis cs:-3,37.509774684906)
--(axis cs:-3,243.962635040283);

\path [draw=color0, semithick]
(axis cs:-2,49.508828163147)
--(axis cs:-2,325.909523010254);

\path [draw=color0, semithick]
(axis cs:-1,77.1266012191772)
--(axis cs:-1,479.225395202637);

\path [draw=color0, semithick]
(axis cs:-0,95.5899562835693)
--(axis cs:-0,590.341850280762);

\addplot [semithick, color0, mark=-, mark size=3, mark options={solid}, only marks]
table {%
-39 5.22390031814575
-38 6.25572681427002
-37 6.56444478034973
-36 6.09748029708862
-35 6.96366310119629
-34 9.30453300476074
-33 8.21129703521729
-32 8.64026260375977
-31 7.84352970123291
-30 8.75040245056152
-29 8.84356427192688
-28 10.8798811435699
-27 11.1100029945374
-26 10.1690502166748
-25 10.3997278213501
-24 12.4184679985046
-23 11.3108296394348
-22 11.7723472118378
-21 11.1935997009277
-20 12.8980417251587
-19 13.4059295654297
-18 11.8703980445862
-17 12.3548753261566
-16 12.2391397953033
-15 14.7252283096313
-14 13.0979306697845
-13 11.6684587001801
-12 11.5907690525055
-11 12.3287258148193
-10 10.5150499343872
-7 15.2682423591614
-6 18.8793020248413
-5 23.5670943260193
-4 30.2904300689697
-3 37.509774684906
-2 49.508828163147
-1 77.1266012191772
-0 95.5899562835693
};
\addplot [semithick, color0, mark=-, mark size=3, mark options={solid}, only marks]
table {%
-39 74.4397230148315
-38 65.9752550125122
-37 80.6355037689209
-36 73.5315656661987
-35 76.6602458953857
-34 84.8390159606934
-33 98.0365562438965
-32 91.4644451141357
-31 92.6722507476807
-30 87.4636917114258
-29 95.9056987762451
-28 110.316257476807
-27 105.531917572021
-26 107.004907608032
-25 106.592775344849
-24 110.657705307007
-23 107.920598983765
-22 111.723009109497
-21 107.256378173828
-20 112.990243911743
-19 113.3626537323
-18 107.675174713135
-17 113.856777191162
-16 116.550710678101
-15 124.263000488281
-14 119.959365844727
-13 113.417181015015
-12 103.946506500244
-11 102.013608932495
-10 98.5654792785645
-7 124.790456771851
-6 142.401453018188
-5 169.25256729126
-4 203.178112030029
-3 243.962635040283
-2 325.909523010254
-1 479.225395202637
-0 590.341850280762
};
\addplot [semithick, color0]
table {%
-7 32.2829475402832
-6 37.5505599975586
-5 43.5465774536133
-4 51.8527145385742
-3 62.4203872680664
-2 77.5186157226562
-1 109.996437072754
-0 131.588882446289
};
\addplot [name path=upper0,draw=none] table{%
	-7 124.790456771851
	-6 142.401453018188
	-5 169.25256729126
	-4 203.178112030029
	-3 243.962635040283
	-2 325.909523010254
	-1 479.225395202637
	-0 590.341850280762
};
\addplot [name path=lower0,draw=none] table{%
	-7 15.2682423591614
	-6 18.8793020248413
	-5 23.5670943260193
	-4 30.2904300689697
	-3 37.509774684906
	-2 49.508828163147
	-1 77.1266012191772
	-0 95.5899562835693
};
\addplot [fill=color0!10] fill between[of=upper0 and lower0];
\end{axis}

\end{tikzpicture}
	\caption{Median attention spread in terms of track age, with $25$th and $75$th percentiles as error bars. \textit{Left:} Track initialization phase, consisting of the first $3.5$ seconds per track. \textit{Right:} Track finalization phase, considering the last $3.5$ seconds per track. Note that only tracks, which reached a final duration of at least $7$ seconds, were included.}
	\label{fig_track_spread}
\end{figure}
\section{CONCLUSION}
In this paper, we investigated the distribution of the attention weight matrices in the transformer decoder in the context of object detection and tracking and proposed a new metric to quantify it. We found that median attention spread decreases with larger IoU, while it increases with larger distance to the ego vehicle. This indicates that attention spread may be able to offer insights into both aleatoric and epistemic uncertainty, whereas it is not possible to differentiate between the two from attention spread alone. Besides this, attention spread observed during the lifetime of tracks changes over time, especially in the initialization and finalization phase.

Our findings open up questions for further research: Is attention spread the best way to describe the attention weight distribution? Can a measure with less variance be found? Can the model design be improved or the model behavior be better understood based on the knowledge about attention spread per layer?

We conclude that the attention matrices available in transformer models have the potential to give insights into detection and tracking uncertainty and that further research is promising.





\bibliographystyle{IEEEtran}

\bibliography{references}

\end{document}